\documentclass[conference]{IEEEtran}
\usepackage{cite}
\usepackage{amsmath,amssymb,amsfonts}
\usepackage{algorithmic}
\usepackage{graphicx}
\usepackage{textcomp}
\usepackage{xcolor}

\def\BibTeX{{\rm B\kern-.05em{\sc i\kern-.025em b}\kern-.08em
    T\kern-.1667em\lower.7ex\hbox{E}\kern-.125emX}}

\begin{document}

\title{MM811 Project Report\\
  \large Cloud Detection and Removal in Satellite Images}
\author{
\IEEEauthorblockN{ Dale Chen-Song, Erfan Khalaji, Vaishali Rani  }
\IEEEauthorblockA{\textit{Department of Computing Science
} \\
\textit{University of Alberta}\\
Edmonton, Canada \\
{\{chensong, khalaji, vrani\}@ualberta.ca}
}}

\maketitle

\begin{abstract}
For satellite images, the presence of clouds presents a problem as clouds obscure more than half to two-thirds of the ground information. This problem causes many issues for reliability in a noise-free environment to communicate data and other applications that need seamless monitoring. Removing the clouds from the images while keeping the background pixels intact can help address the mentioned issues. Recently, deep learning methods have become popular for researching cloud removal by demonstrating promising results, among which Generative Adversarial Networks (GAN) have shown considerably better performance. In this project, we aim to address cloud removal from satellite images using AttentionGAN and then compare our results by reproducing the results obtained using traditional GANs and auto-encoders. We use RICE dataset. The outcome of this project can be used to develop applications that require cloud-free satellite images. Moreover, our results could be helpful for making further research improvements.

\end{abstract}


\section{Introduction}
Cloud coverage on satellite images makes it challenging to use such data for applications that require seamless input. With recent advances in machine learning, generative adversarial networks (GANs) have shown promising results in reconstructing satellite images without cloud coverage. While the importance of GANs is highlighted in recent studies, the lack of implementation for real-world applications remains unaddressed. StyleGAN is an extension of GAN architecture, which proposes large changes to the generator model. In this project, we aim to use the dataset introduced in~\cite{sentinel2ebel} to train StyleGANs and compare our results by reproducing the obtained precision in~\cite{sentinel2ebel,lishenclouddetection,zy3chen}. Further, we aim to train StyleGANs by modifying loss function, network structure, and hyperparameters. The purpose of doing so is to complete our research and explore different network configurations. Comparing our obtained results with~\cite{sentinel2ebel,lishenclouddetection,zy3chen} is a significant contribution to understanding how StyleGANs, GANs, and auto-encoders behave differently while being fed the same satellite data. The novelty of our project is centered around two points. Firstly, to the best of our knowledge, we are the first research group that is training StyleGANs with the dataset introduced in~\cite{sentinel2ebel}. Secondly, auto-encoders are not trained with the dataset that ~\cite{sentinel2ebel} published.

\section{Brief Summary of what exists}
\subsection{Convolution Neural Network}
 \cite{zy3chen} used a dataset of images from ZiYuan-3 (ZY-3) high resolution satellite images~\cite{zy3chen}. For their cloud detection, they used a convolution neural network (CNN) network due to the ZY-3 satellite images only having three visible bands and on the near-infrared band, posing a challenge. \cite{zy3chen} used traditional CNN, consisting of stacking multiple players of convolution and pooling in alternating fashion, but changing the fully connected layer with global average pooling (GAP). GAP’s advantage is that it is more native to convolution structure, improving the cloud detection CNN architecture. 
For cloud removal, they proposed using content-texture-spectral CNN (CTS-CNN) to recover areas that were covered by clouds. CTS-CNN can remove the thick and thin clouds and cloud shadows. Content generation networks were used to reconstruct the missing objects in the image with clouds. The spectral generation network restores the spectral information of the missing objects. A texture generation network is used to enhance the content of the content network. Rectified linear units were applied in every output of the convolutional layer. Images with clouds were used as input data and images without clouds were used as targets to recover the region with clouds. While this proposed algorithm performs decently with reconstructing image information under clouds with the ZY-3 images, it still cannot be used in cases where land cover may change a lot.

\subsection{Multi-Scale Convolutional Feature Fusion}
Li et al. used multi-scale convolutional feature fusion (MSCFF) for remote sensing images of different sensors \cite{lishenclouddetection}. It uses a symmetric encoder-decoder module to extract multi-scale and high-level spatial features. Cloud and cloud shadow detection is a challenge that needed to be resolved by the algorithm. 
Rule-based classification technique is used for extracting clouds from images. In the case of images having limited spectral information, bright non-cloud objects are delegated and thin clouds are omitted. 
o overcome this challenge, they try to create masks for multiple types of satellite images. MSCFF model consists of a symmetric encoder decoder module and a multi-scale feature fusion module. In the training phase, the model is made to learn iteratively and we receive cloud and cloud shadow maps as output features maps of MSCFF. In the testing phase, pre-processed test images are fed into a pre-trained MSCFF model which will predict cloud and shadow for each pixel of the image. The output map of the feature will then be fed into a binary classifier outside the model for pixel-wise binarization. The output of the previous step will yield us binary cloud and shadow mask combined together. 
The same dataset which was used to train MSCFF was used to train the Deeplab and DCN network. In order to adjust the inputs from multiple channels ~\cite{lishenclouddetection}. They changed the first layer of both models i.e Deeplaband DCN. If one pixel is expanded, it can boost the performance in accuracy assessment for cloud shadow detection results for Fmask whereas the result of MSCFF remains constant.
On checking the efficiency of Fmask, Deeplab, DCN, and MSCFF, the outcome that they can infer from the implementation of the above-mentioned algorithms is that deep learning-based algorithms take less time than the traditional Fmask technique.

Some limitation is that, even though Li et al. can achieve higher accurate results using MSCFF, the architecture of MSCFF has restrictions for the size of the input image that can be processed on. Also, MSCFF can identify it as the wrong class. For example, snow can be classified as clouds because central areas for bright surfaces are identical and deep learning models cannot pull out discriminative features in the patch-wise processing.

\subsection{Generative Adversarial Network}
GANs are a generative model that is an unsupervised learning task in machine learning that can determine patterns and regularities from the input data to generate new output that could be part of the original dataset. GANs use two neural networks, a generator to train generate new examples, against a discriminator model that classifies the examples from the generator as real or fake, until the discriminator model is fooled more than half the time. GANs were used to generate cloud-free images such as done by Ebel et al. \cite{sentinel2ebel}.

\subsection {Style Generative Adversarial Network}
StyleGANs were proposed for the first time in 2014. Their purpose is to synthesize artificial examples, such as pictures that are obscure from authentic photographs. A typical example of a GAN application is to produce artificial face pictures by learning from a dataset of notable faces. However, in this project, we aim to use StyleGANs to reconstruct cloudy images. Given that we have access to the segmentations (pixel labels) and the original images, we are aiming to modify style GANs in such a way that they are capable of reconstructing the image without the clouds. With respect to the fact that style GANs perform significantly better than traditional GANs, we are hoping to outperform the results mentioned in~\cite{sentinel2ebel}.

\subsection{Auto-Encoder}
Auto-encoders are an unsupervised learning technique, that leverages neural networks for the task of representation learning. They are a specific type of feedforward neural network where the input is the same as the output. It composes of 3 components, the encoder, which compresses the input, the code, which is produced by the encoder, and the decoder, which then reconstructs the input only using the code. They are used as another approach for image reconstruction. Denoising auto-encoders, in particular, are used for outlier detection, and in general, detecting noise in a search space. While auto-encoders are computationally less expensive than GAN models, their promising performance has led them to be used in various computing applications. Therefore, we see it as a crucial step to have it as a part of our benchmark for model evaluation.

\section{Milestones and Breakdown}
This is the general milestones and breakdown of tasks. 
    
    \begin{itemize}
    \item Literature Review: All, October 28
    \item Project: November 30
    \begin{itemize}
    \item Creating the Pipeline: All, November 10
    \item Training GAN: Dale, November 21
    \item Training StyleGAN: Erfan, November 21
    \item Training Auto-Encoder: Vaishali, November 21
    \end{itemize}
    \item Project Demo: All, December 5
    \item Final Report Draft: All, December 7
    \item Project Implementation: All, December 17
    \item Final Report: All, December 17
    \end{itemize}

\section{Literature Review}
Cloud removal is a matter of extracting the background, and in general, background extraction has various applications in industry and academic research. Furthermore, there are different techniques for background removal, each of which might be suitable for a particular application. Authors in~\cite{yang2017background} introduce an approach titled Stability of Adaptive Features (SoAF) that adaptively weighs the contribution of pixels by utilizing the stability of different features to extract the image foreground. By computing local invariances,~\cite{yang2017background} aims to tackle the issue of object sensitivity to shadows. To this end, the authors improved the SILTP algorithm dominating mechanism in addressing the mentioned issue. The results indicate that the algorithm works with high performance for office-related, pets, cubicles, and other datasets. ~\cite{ma2019background} proposes a method that works based on optical extraction. In this study, after extracting the optical features of an image, authors perform image decomposition through RPCA, and after that, they use the integration of the foreground mask to obtain their final result. For model evaluation, precision and recall were used, and they are among the commonly used metrics. Looking at the reported numbers and the visuals, incorporating RPCA with the models' authors of ~\cite{ma2019background} proposed looks promising.

In~\cite{ge2020deep}, authors propose a method for tackling this challenge. The core idea behind this study is to consider pixel variation rather than pixel distribution when it comes to foreground detection. Foreground detection is a binary classification problem, and~\cite{ge2020deep} uses variation transformation to address it. For variation representation,~\cite{ge2020deep} forms patches for every image pixel so that spatial information can be preserved and assumed to be beneficial for this method. On top of variation transformation, they run a mechanism called Deep Variation Transformation Network (DVTN). DVTN tries to learn pixel variations using the processed data in the previous stage. Then, The temporal information obtained by the network is used for foreground detection that shows promising results. Given the recent advancements in the foreground and background detection, it would be beneficial to incorporate such progress with cloud removal techniques. In GLF-CR, there are two streams in which SAR features are hierarchically fused into optical features for the sake of restoring the lost information~\cite{xu2022glf}. At the top of the network, there are shallow feature extraction and cloud-free image reconstruction networks which then pass their outputs to a network that consists of convolution layers with a Relu activation function. Convolution layers are integrated into the network to learn the pixel values that are at the end passed to an image reconstruction network. The authors state that they obtain slightly higher SSIM compared to the state-of-the-art method.

In addition to convolution networks, residual networks also proved to be useful for classification problems. One of the recent and most cited publications in the field of cloud removal uses residual networks to address this issue~\cite{meraner2020cloud}. The aim of this model is to learn the pixel values of the parts of the cloud-free images so that it can reconstruct the cloudy images by adding all the learned features at the bottom of the network. Even though the structure of the network introduced in~\cite{meraner2020cloud} is not as complex as the one in~\cite{xu2022glf}, it demonstrates better results. Authors in~\cite{mahalingaiah2019semantic} introduce a novel methodology for semantic learning in images. They modified VGG19 as a part of their model and define regions of interest for processing the images. According to their result, Resnet50 has a less fluctuating learning process. This model obtains 17.4 compression ratio while maintaining a score of 0.85 SSIM. Even though the model does not perform the state-of-the-art, the novelty in designing a simple network given the obtained scores shows that a complex network is not necessarily the best, and image semantic learning is possible through designing a simple network.

Generative Adversarial Networks (GAN) are relatively newer deep learning methods widely used for image translation, reconstruction, and transformation. One of the recent studies uses Conditional GANs (cGAN) to learn the non-linear mapping of two images in two different domains~\cite{bermudez2018sar}. This study first uses image reconstruction to generate images closer to real images. After having enough reconstructed data samples, in the next phase of the network, both real images and reconstructed images are given to a discriminator. The loss is computed and propagated to the beginning of the network, where the image generator will have the chance to use the feedback to reconstruct more realistic images. After learning the optical features through cGAN, these features are used to reconstruct the cloudy SAR images. Using cGANs leverages this study by helping to obtain more accurate results. Alternatively,~\cite{pan2020cloud} uses Spatial Attention GANs (SpaGANs) to simulate a human visual system where humankind focuses on the cloudy part to collect as much as visual information as they could. This method enhances the possibility of reconstructing the image in a more efficient and accurate manner. The authors of ~\cite{pan2020cloud} used spatial attentive blocks as well as residual entities beneath a convolutional layer in which the result is activated using a Relu function.~\cite{pan2020cloud} uses SSIM to compare their work with cycle GANs and conditional GANs. The SSIM scores demonstrate that the SpaGANs perform better compared to the cycle/conditional GANs.

Unlike~\cite{bermudez2018sar} that uses RBG satellite images to train cGANs with them, authors of~\cite{enomoto2017filmy} use an extended version of RBG images in the forms of multi-spectral images. Therefore, they turn RBG images to have four channels as a part of their data preparation pipeline. The network architecture of this study consists of convolution layers. Relu activation function, batch normalization, and drop-out functionalities. In the Relu layers, authors used leaky Relu for both encoders and discriminators. In this research paper, the visual results seem to be clear. however, they have no table or clear methodology to make the readers able to compare the results with other studies. With an image-to-image translation method,~\cite{gao2020cloud} uses a hybrid two-stage network where CNN and GANs work together to accomplish cloud removal using SAR data. In the first step, a specially designed convolutional neural network (CNN) translates the synthetic aperture radar (SAR) images into simulated optical images in an object-to-object manner; in the second step, the simulated optical image, together with the SAR image and the optical image corrupted by clouds, is fused to reconstruct the corrupted area by a generative adversarial network (GAN) with a particular loss function. The proposed model achieves an SSMI score of over 0.9 which makes it more accurate compared to the other approaches that exist in the body of the literature.

One of the more popular methods to solve this is synthetic aperture radar (SAR)-based methods as SAR imaging isn't affected by clouds, and can reflect ground information differences and changes. \cite{chen_sar} proposed a cloud removal method using SAR-optical data fusion and graph-based feature aggregation network (G-FAN). Cloudy optical images and SAR images are concatenated and transformed into hyper-feature maps, which are then inputted into G-FAN to obtain the missing data of the cloud-covered areas, as it aggregates the information from the SAR image and spectral information of neighbourhood and non-neighbourhood pixels of optical images. A loss function based on smooth L1 loss function and Multi-Scale Structural Similarity Index is used for the model. This proved to show better results than the traditional deep learning model. 

\cite{wang_simulated_radiance} on the other hand proposed an alternative way of detecting and removing clouds that are less complicated for a common user to use, by using simulated top-of-atmosphere radiance fields. Based on these fields, a simple scheme is provided to reconstruct the cloud-contaminated images based on near-infrared and visible bands. Then, to recover the band radiance of the contaminated pixels based on simulated radiance fields. This proposed method can be used in common satellite channels.

For example, \cite{zhaobackground_cues} describes a method for background subtraction by determining foreground and background cues, differing from the norm of improving the accuracy of motion estimation. The framework they created was the Integration of Foreground and Background (IFB). Extracting the foreground cues was done by using a Gaussian mixture model. Background cues were obtained from finding the static spatiotemporal features filtered by homography transformation. The integration of the alternate cues was determined by super-pixels.

\cite{zhao_dpdl} proposed another method for background subtraction using Deep Pixel Distribution Learning (DPDL), which uses Random Permutation of Temporal Pixels (RPoTP) to represent the distribution of past observations for a particular pixel. By looking at a variety of pixel values, they can determine whether the pixel varies due to illumination or a breeze for example, as it follows a pattern that is detected. Then, a CNN is used to learn if the pixel is in the foreground or background. \cite{zhao_dynamic_dpdl} further refined their method by proposing a dynamic DPDL by dynamically generating RPoTP features to prevent overfitting. A Bayesian refinement model was used to handle the random noise that was generated from the random permutation. Another approach for background subtraction from \cite{zhao_background_ADNN} was a framework for Arithmetic Distribution Neural Network, which can learn the distribution of temporal pixels. Using an improved Bayesian refinement model based on neighbouring information

The Smart MultiMedia conference is a forum where it promotes the exchange of the latest in multimedia, whether it is technologies, systems, and applications in research, development, or industrial perspectives. Authors are invited to submit papers for the conference. Here are some of the papers we reviewed from the conference in 2019. 

A method that \cite{wu_background_clustering} proposed using clustering to differentiate the background and foreground pixels, using the difference vectors between pixels in the background image set and the current frame, saving computational time by only iterating once. The clustering method used was the quartile method, but only needed 2 clusters as it is just the background and foreground clusters. This method saves a lot of time compared to other background subtraction methods while providing similar results.

\cite{liu_fog_removal} worked on removing fog from images to improve pilots' perception of their safety in take-off and landing situations. Image defogging removes undesirable factors from the image, such as weather and blurring. The authors proposed a method to defog the image by correcting the original image with Gamma as the guided map to correct contrast and brightness. The gamma-corrected images are then put through a guided filter, which then was put through a Retinex model to defog the image. The images are then combined with a histogram truncation technique to output the mapping between 0 and 255. The algorithm proposed in this paper removed the fog, improve the contrast and maintain consistency in colour. ~\cite{ke_iris} proposed a method to classify the race based on iris image segmentation using machine learning. Iris recognition can be used in many various fields of identification, as iris patterns are unique. \cite{ke_iris} first used feature extraction and classification, then used iris image segmentation. For feature extraction and classification, they built a classifer, local Gabor binary pattern, and support vector machine. This was used to separate the irises into 2 groups, human vs non-human. Afterward, they use Hough transform to segment the interior boundary and localize the exterior boundary by using localizing region-based active contour model.

Deep learning is proposed for cleaning and imputing satellite images. In this approach, supplementary data are used to obtain a coherent and cloud-free image for a specific target date and region using a modified version of the well-known U-Net architecture~\cite{oehmcke2020creating}. Clouds can obscure or cover the land surface, making it difficult to analyze the data. The FMASK approach was developed to detect distortions in satellite imagery, but it is still far from perfect. We propose a model based on deep learning that can efficiently handle large amounts of satellite time series data instances. The time series data are aggregated via a special gating function, and the U-Net architecture is used to generate a cloud- and distortion-free image. This approach outperforms the commonly used baseline approach.

 ~\cite{chen2019thick} proposes a cloud detection and removal method based on deep learning architecture. The problem statement is addressed by dividing it into two sub-problems which are a convolution neural network (CNN) architecture used for cloud detection and also used CNN. However, multi-source data is used (content, texture, and spectral) as an input of the unified framework. All being said, the proposed algorithm cannot be applied to cases in which the land cover may significantly change. In another study,~\cite{ebel2020multisensor} proposes a method based on using YUV colour space to reconstruct cloud-free images, and a residual symmetrical encoding-decoding architecture was used to recover detailed information without downsampling and upsampling layers. By using deep convolutional neural networks (CNNs), CR-GAN-PM decomposes cloud-free background layers and cloud distortion layers. In this paper, we introduce a YUV-GAN, a residual encoding-decoding network, and a fidelity loss in YUV colour space for thin cloud removal on satellite imagery. The experimental results prove that the adversarial learning between the generator and residual discriminator can improve the accuracy of ground scene identification. Authors of~\cite{borkiewicz2021cloudfindr} describe a method for creating cloud artifact masks from satellite imagery by combining traditional image processing with deep learning based on U-Net, in this paper. The underlying data is a 1-dimensional elevation model, and the general goal is to visualize scientific data in a cinematic manner. To create realistic landscapes and backdrops in computer-generated movies, CloudFindr masks cloud artifacts in satellite-collected DEM data. A popular U-Net algorithm is used in this paper to detect and mask out cloud regions using deep-learning image segmentation. CloudFindr is a method for labeling pixels as "cloud" or "non-cloud" from a single-channel DEM image. It has high accuracy.
 
 In another study,~\cite{saxena2022deep} proposes a deep-learning neural network that is applied to the reconstruction of hazy satellite images with a MSE of 0.125 and a PSNR of 80. During rainy seasons, the proposed model does not require pre-training datasets. The deep learning model was found to be highly effective, efficient, adaptive, and can be applied in all environments as a means of detecting and removing clouds from images. We only have rectangular masks to apply, and the MSE value is quite acceptable, but some hyperparameters and parameter tuning are needed to produce better results. Similarly, in another study concerned with landscape image reconstruction (\cite{langer2020formatting}), a generative adversarial network is used to generate satellite imagery conditioned on population changes, also known as the latent space, by generating random noises. Using an Adversarial Latent AutoEncoder (ALAE) model trained on satellite data and population data, we generated and edited images via a trained generative model. They use ALAE for the basis of our model, which is in turn based on StyleGAN.
 
 Given that they could think of addressing cloud removal with unsupervised methods, ~\cite{de2019unsupervised} proposes an efficient unsupervised method for detecting relevant changes between two temporally different images of the same scene. The method uses a convolutional neural network for semantic segmentation. It is necessary to train a convolutional neural network at various levels of a trained CNN in order to create an effective DI. The CNN consists of an encoder and a decoder, which together extract features from the input image and build a segmented representation. Using PCC2, the semantic segmentation results were evaluated, and the best results were obtained at epoch 15 when 89.2\% accuracy was achieved.

\section{Methods}

The proposed approach for this project is using AttentionGAN, an architecture shown below in Fig.~\ref{aggan}. The novelty in this project is by using GANs while preserving cycle consistency, computing loss for clouds to clear, computing loss for clear to clouds and also preserving attention which leads to preserving background pixels through computing pixel loss and shifting attention towards the changes in images (clouds, shadows, etc. of images).
\begin{figure*}
\centering
\includegraphics[width=\textwidth,height=7cm]{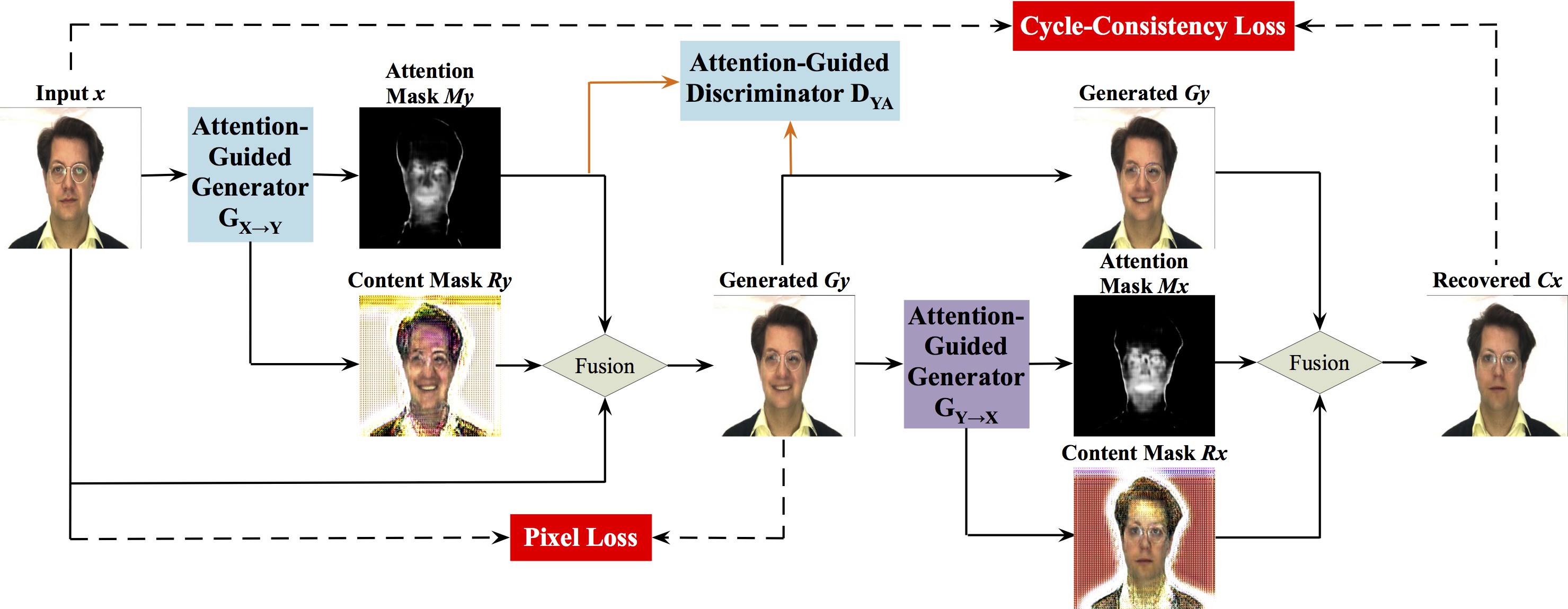}
\caption{\label{aggan} \textbf{Architecture of AttentionGAN} The attention-guided generators
have built-in attention mechanisms, which can detect the most discriminative part of images. After that, the input image, content mask, and attention mask are fused together to synthesize the targeted image. Moreover, an attention-guided discriminator Dya is added to distinguish only the most discriminative content}
\end{figure*}

\subsection{Attention Guided Generator}
GANs are composed of two competing modules; the generator G and the discriminator D. These two neural networks contest with each other in the form of a zero-sum game, where one's gain is the other's loss. X denotes training images in the source and Y denotes the target image domain. Therefore, the  generator Gx->y maps x from the source domain to the generated image Gy in the target domain Y and tries to trick the discriminator Dy. Discriminator Dy will improve itself so that it can determine whether the sample is a generated or a real photo. This is similar to Gy->x and Dx.

There are two mappings between X and Y via two generators with built-in attention mechanisms. Gx->y:x->[My, Ry, Gy] and Gy->x:y->[Mx, Rx, Gx], where M is the mask, R is the content mask, and G is generated images. The attention masks M define a per-pixel intensity specifying to which extend each pixel of the content masks R will contribute to the final rendered image. After that, the input image x, the generated attention mask My, and the content mask Ry are fused to generate the target image Gy. 

\subsection{Attention Guided Discriminator}
The generators only act on the attended regions. However, the basic discriminators only consider the whole image currently. The discriminator Dy takes the generated image Gy or the real image y as input and tries to distinguish them, while the discriminator Dx takes the generated image Gx or the real image x. Therefore, the attention-guided discriminator would be similar to the basic discriminator but will take the attention mask as input. Therefore, the attention-guided discriminator Dya tries to distinguish the fake image pairs, My and Gy, and the real image pairs, My and y, while Dxa tries to distinguish the fake image pairs, Mx and Gx, and the real image pairs, Mx and x. 

\subsection{Loss function}
The loss function that we wanted to initially implement was the Tversky index loss function~\cite{lossfunction}. However, even though this was used as it provided more generalization and better optimization with an imbalanced dataset, and this proved to work effectively with attention U-Net, we had to use other loss functions in our method. AttentionGAN has various loss functions such as pixel loss for both generator and discriminator, given that we have two generators and two discriminators, the number of loss functions multiplies. In our model, we optimize two generator and two discriminator loss functions. Last but not least, we optimize a consistency loss function mostly because AttentionGAN has a cycle consistency attribute that we had to attend to.

\subsection{Dataset}
Remote sensing Image Cloud rEmoving (RICE) ~\cite{lin2019remote} dataset contains two parts: RICE1 contains 500 pairs of images, each pair has images of cloud and cloudless lands with the size of 512 x 512 without overlap; RICE2 contains 450 sets of images, each set contains three 512 x 512 size images, respectively, the reference picture without clouds, the picture of the cloud and the mask of its cloud. RICE1 image set was used in this project as RICE1 contained thin clouds, which made for easier computation for training the model, while RICE2 has mask associated and images having thick clouds. The dataset is available at: https://github.com/BUPTLdy/RICE\textunderscore DATASET
The image dataset was separated so that 450 images were used for training, and 50 images were used for testing. 

\subsection{Parameter Settings}

We also performed data augmentation by rotating, flipping (vertically and horizontally), cropping, and changing the background color. As a result, we managed to change the number of data values from 450 to over 1600 images. In order to avoid the vanishing gradient issue and the possibility of our optimization not diverging, we decided to use an epoch decay. In other words, we gradually decrease the epochs until the model diverges and the training process ends. In this project, we used a different number of epochs. We initially started with three epochs and continued to 30 and 300. However, the results we obtained for 300 were not very different than 30 epochs so we ruled it out. The last parameter we adjusted was the batch size. Since we did not have a great volume of data points, we decided to keep the batch size at 1. Therefore, for each epoch, every single image would be fed into the model individually until all images in the training folder go through the same process.

\section{Results}
According to the available computational resources we had to complete this project, we decided to train AttentionGAN based on three different epoch values. In the first stage, we trained the model with only 3 epochs to understand how well the model will perform on the test set, therefore it was not trained for a long period of time. In order to demonstrate our results, we chose three photos each with a different texture background.

Figure 2 demonstrates a cloudy image of land without any plant/forest cover, ocean, and land with plant cover. Furthermore, Figure 2 demonstrates the same images generated after 3 and 30 epochs using AttentionGAN. Looking at the image and analyzing it qualitatively illustrates that the background pixel values are distorted and this is not a desirable output. Therefore, we increased the epochs to 30 to let the model train more. After 30 epochs background pixels are matching the reference image. Further, the thin cloud is less in the image compared to the cloudy reference image. This can be qualitatively assessed by comparing the background color intensity. It is worth noting that after 30 epochs, AttentionGAN shows a promising result in preserving background pixels.

\begin{figure*}
\centering 
\includegraphics[width=13cm,height=7cm]{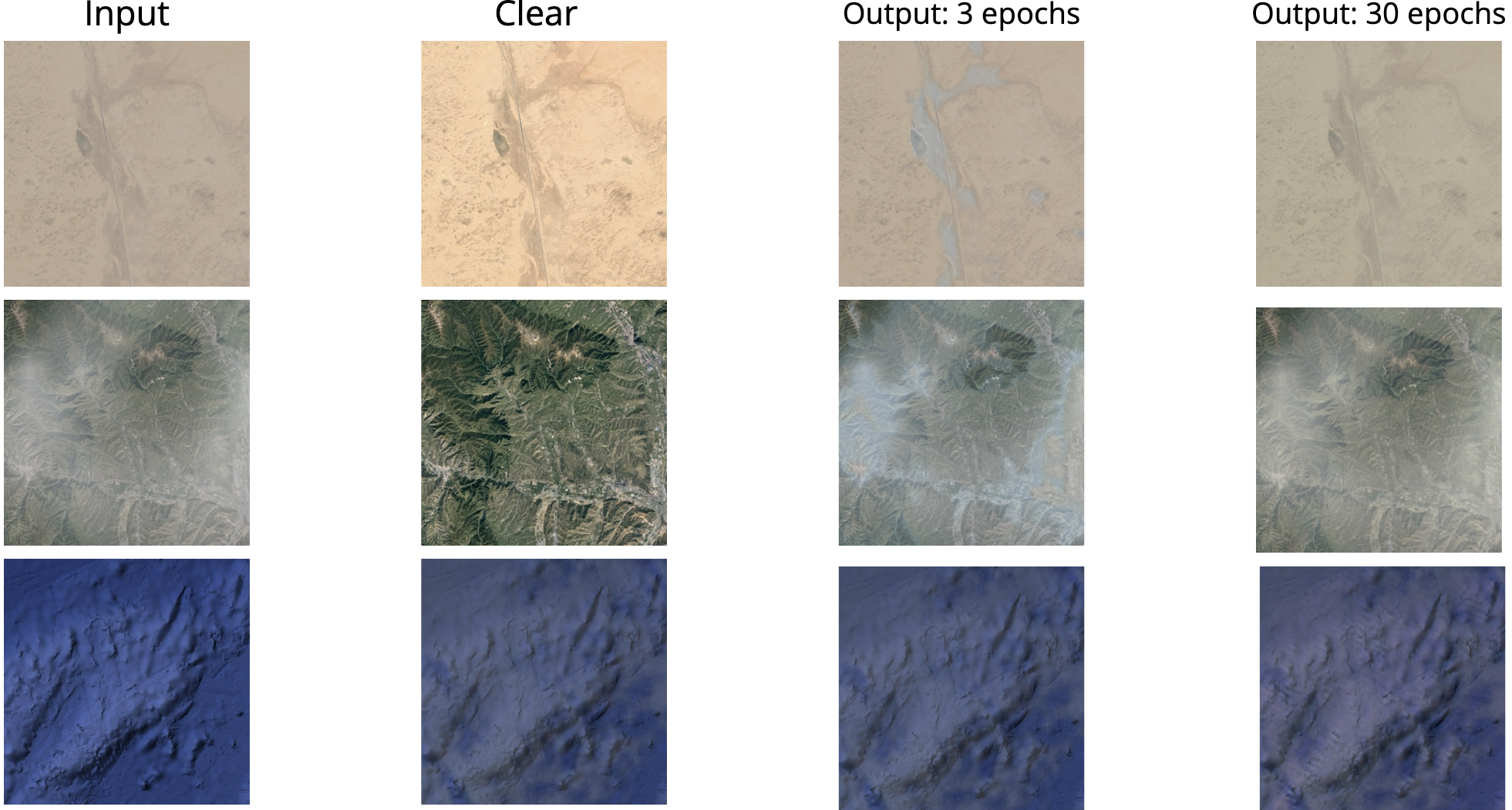}
\caption{Versions of satellite images with clouds, then no clouds, then 3 epochs through AttentionGAN, and then 30 epochs through AttentionGAN}
\end{figure*}

% \begin{figure}
% \centering 
% \includegraphics[width=5cm,height=5cm]{Images/0001_real_Clear.png}
% \caption{Clear version of a satellite image from a land without a forest cover}
% \end{figure}

% \begin{figure}
% \centering 
% \includegraphics[width=5cm,height=5cm]{Images/3_fake_Clear_1.png}
% \caption{Clear version of a satellite image from a land without a forest cover after 3 epochs}
% \end{figure}

% \begin{figure}
% \centering 
% \includegraphics[width=5cm,height=5cm]{Images/0030_fake_Clear_30.png}
% \caption{Clear version of a satellite image from a land without a forest cover after 30 epochs}
% \end{figure}

% \begin{figure}
% \centering 
% \includegraphics[width=5cm,height=5cm]{Images/0001_real_Clear.png}
% \caption{Clear version of a satellite image from a land without a forest cover}
% \end{figure}

\section{Discussion and Conclusion}
We proposed using AttentionGAN to generate cloudless images from cloudy images. Even though we could only partially remove the thin cloud in RICE1 dataset, our approach demonstrates promising results in preserving background pixel values. Moreover, to the best of our knowledge, this study is the first implementation of AttentionGAN adapted for cloud removal, which can be considered a considerable contribution to the body of the literature for the reasons we describe.

Firstly, we provide empirical evidence that AttentionGAN shows considerably good results in preserving background images when we are concerned with removing foreground pixels. Secondly, we show that AttentionGAN, only after 3 epochs, reaches a point where it can distinguish foreground and background depending on the provided data. Thirdly, our results demonstrate that AttentionGAN is potentially a useful method for cloud removal from satellite images. And last but not least, we highlight the importance of two concepts in GANs; Cycle Consistency and Attention. These two features of our model show that the model can be useful in removing clouds from satellite images.

It goes without saying that we experienced limitations in this study. GANs are extremely data-hungry and we managed to perform data augmentation and change the number of data points from 450 to 1600 by rotating, flipping, cropping, and changing the background color of images. Also, we would have tested the model for 3000 epochs if we had access to more data and computational resources. Furthermore, the clouds were not fully removed from the images.

As a future work, it would be an asset to try training AttentionGAN on a dataset with a greater volume of images using better graphical processing units. Also, one potential approach would be using attention-U-Net, or only U-Net as a replacement for the generator in AttentionGAN model. However, we would look to fully remove the clouds with the next model.

\bibliographystyle{IEEEtran}
\bibliography{access.bbl}

\begin{thebibliography}{10}
\providecommand{\url}[1]{#1}
\csname url@samestyle\endcsname
\providecommand{\newblock}{\relax}
\providecommand{\bibinfo}[2]{#2}
\providecommand{\BIBentrySTDinterwordspacing}{\spaceskip=0pt\relax}
\providecommand{\BIBentryALTinterwordstretchfactor}{4}
\providecommand{\BIBentryALTinterwordspacing}{\spaceskip=\fontdimen2\font plus
\BIBentryALTinterwordstretchfactor\fontdimen3\font minus
  \fontdimen4\font\relax}
\providecommand{\BIBforeignlanguage}[2]{{%
\expandafter\ifx\csname l@#1\endcsname\relax
\typeout{** WARNING: IEEEtran.bst: No hyphenation pattern has been}%
\typeout{** loaded for the language `#1'. Using the pattern for}%
\typeout{** the default language instead.}%
\else
\language=\csname l@#1\endcsname
\fi
#2}}
\providecommand{\BIBdecl}{\relax}
\BIBdecl

\bibitem{sentinel2ebel}
P.~Ebel, A.~Meraner, M.~Schmitt, and X.~X. Zhu, ``Multisensor data fusion for
  cloud removal in global and all-season sentinel-2 imagery,'' \emph{IEEE
  Transactions on Geoscience and Remote Sensing}, vol.~59, no.~7, pp.
  5866--5878, 2021.

\bibitem{lishenclouddetection}
\BIBentryALTinterwordspacing
Z.~Li, H.~Shen, Q.~Cheng, Y.~Liu, S.~You, and Z.~He, ``Deep learning based
  cloud detection for medium and high resolution remote sensing images of
  different sensors,'' \emph{ISPRS Journal of Photogrammetry and Remote
  Sensing}, vol. 150, pp. 197--212, 2019. [Online]. Available:
  \url{https://www.sciencedirect.com/science/article/pii/S0924271619300565}
\BIBentrySTDinterwordspacing

\bibitem{zy3chen}
Y.~Chen, L.~Tang, X.~Yang, R.~Fan, M.~Bilal, and Q.~Li, ``Thick clouds removal
  from multitemporal zy-3 satellite images using deep learning,'' \emph{IEEE
  Journal of Selected Topics in Applied Earth Observations and Remote Sensing},
  vol.~13, pp. 143--153, 2020.

\bibitem{yang2017background}
D.~Yang, C.~Zhao, X.~Zhang, and S.~Huang, ``Background modeling by stability of
  adaptive features in complex scenes,'' \emph{IEEE Transactions on Image
  Processing}, vol.~27, no.~3, pp. 1112--1125, 2017.

\bibitem{ma2019background}
Y.~Ma, G.~Dong, C.~Zhao, A.~Basu, and Z.~Wu, ``Background subtraction based on
  principal motion for a freely moving camera,'' in \emph{International
  Conference on Smart Multimedia}.\hskip 1em plus 0.5em minus 0.4em\relax
  Springer, 2019, pp. 67--78.

\bibitem{ge2020deep}
Y.~Ge, J.~Zhang, X.~Ren, C.~Zhao, J.~Yang, and A.~Basu, ``Deep variation
  transformation network for foreground detection,'' \emph{IEEE Transactions on
  Circuits and Systems for Video Technology}, vol.~31, no.~9, pp. 3544--3558,
  2020.

\bibitem{xu2022glf}
F.~Xu, Y.~Shi, P.~Ebel, L.~Yu, G.-S. Xia, W.~Yang, and X.~X. Zhu, ``Glf-cr:
  Sar-enhanced cloud removal with global--local fusion,'' \emph{ISPRS Journal
  of Photogrammetry and Remote Sensing}, vol. 192, pp. 268--278, 2022.

\bibitem{meraner2020cloud}
A.~Meraner, P.~Ebel, X.~X. Zhu, and M.~Schmitt, ``Cloud removal in sentinel-2
  imagery using a deep residual neural network and sar-optical data fusion,''
  \emph{ISPRS Journal of Photogrammetry and Remote Sensing}, vol. 166, pp.
  333--346, 2020.

\bibitem{mahalingaiah2019semantic}
K.~Mahalingaiah, H.~Sharma, P.~Kaplish, and I.~Cheng, ``Semantic learning for
  image compression (slic),'' in \emph{International Conference on Smart
  Multimedia}.\hskip 1em plus 0.5em minus 0.4em\relax Springer, 2019, pp.
  57--66.

\bibitem{bermudez2018sar}
J.~Bermudez, P.~Happ, D.~Oliveira, and R.~Feitosa, ``Sar to optical image
  synthesis for cloud removal with generative adversarial networks.''
  \emph{ISPRS Annals of Photogrammetry, Remote Sensing \& Spatial Information
  Sciences}, vol.~4, no.~1, 2018.

\bibitem{pan2020cloud}
H.~Pan, ``Cloud removal for remote sensing imagery via spatial attention
  generative adversarial network,'' \emph{arXiv preprint arXiv:2009.13015},
  2020.

\bibitem{enomoto2017filmy}
K.~Enomoto, K.~Sakurada, W.~Wang, H.~Fukui, M.~Matsuoka, R.~Nakamura, and
  N.~Kawaguchi, ``Filmy cloud removal on satellite imagery with multispectral
  conditional generative adversarial nets,'' in \emph{Proceedings of the IEEE
  Conference on Computer Vision and Pattern Recognition Workshops}, 2017, pp.
  48--56.

\bibitem{gao2020cloud}
J.~Gao, Q.~Yuan, J.~Li, H.~Zhang, and X.~Su, ``Cloud removal with fusion of
  high resolution optical and sar images using generative adversarial
  networks,'' \emph{Remote Sensing}, vol.~12, no.~1, p. 191, 2020.

\bibitem{chen_sar}
\BIBentryALTinterwordspacing
S.~Chen, W.~Zhang, Z.~Li, Y.~Wang, and B.~Zhang, ``Cloud removal with
  sar-optical data fusion and graph-based feature aggregation network,''
  \emph{Remote Sensing}, vol.~14, no.~14, 2022. [Online]. Available:
  \url{https://www.mdpi.com/2072-4292/14/14/3374}
\BIBentrySTDinterwordspacing

\bibitem{wang_simulated_radiance}
T.~Wang, J.~Shi, H.~Letu, Y.~Ma, X.~Li, and Y.~Zheng, ``Detection and removal
  of clouds and associated shadows in satellite imagery based on simulated
  radiance fields,'' \emph{Journal of Geophysical Research: Atmospheres}, vol.
  124, 07 2019.

\bibitem{zhaobackground_cues}
C.~Zhao, A.~Sain, Y.~Qu, Y.~Ge, and H.~Hu, ``Background subtraction based on
  integration of alternative cues in freely moving camera,'' \emph{IEEE
  Transactions on Circuits and Systems for Video Technology}, vol.~29, no.~7,
  pp. 1933--1945, 2019.

\bibitem{zhao_dpdl}
C.~Zhao, T.-L. Cham, X.~Ren, J.~Cai, and H.~Zhu, ``Background subtraction based
  on deep pixel distribution learning,'' in \emph{2018 IEEE International
  Conference on Multimedia and Expo (ICME)}, 2018, pp. 1--6.

\bibitem{zhao_dynamic_dpdl}
C.~Zhao and A.~Basu, ``Dynamic deep pixel distribution learning for background
  subtraction,'' \emph{IEEE Transactions on Circuits and Systems for Video
  Technology}, vol.~30, no.~11, pp. 4192--4206, 2020.

\bibitem{zhao_background_ADNN}
C.~Zhao, K.~Hu, and A.~Basu, ``Universal background subtraction based on
  arithmetic distribution neural network,'' \emph{IEEE Transactions on Image
  Processing}, vol.~31, pp. 2934--2949, 2022.

\bibitem{wu_background_clustering}
X.~Wu, X.~Gao, C.~Zhao, J.~Wu, and A.~Basu, ``Background subtraction by
  difference clustering,'' in \emph{Smart Multimedia}, T.~McDaniel,
  S.~Berretti, I.~D.~D. Curcio, and A.~Basu, Eds.\hskip 1em plus 0.5em minus
  0.4em\relax Cham: Springer International Publishing, 2020, pp. 45--56.

\bibitem{liu_fog_removal}
X.~Liu, C.~Liu, and H.~Lan, ``Fog removal of aerial image based on gamma
  correction and guided filtering,'' \emph{Smart Multimedia}, pp. 472--479,
  2020.

\bibitem{ke_iris}
X.~Ke, L.~An, Q.~Pei, and X.~Wang, ``Race classification based iris image
  segmentation,'' in \emph{Smart Multimedia}, T.~McDaniel, S.~Berretti,
  I.~D.~D. Curcio, and A.~Basu, Eds.\hskip 1em plus 0.5em minus 0.4em\relax
  Cham: Springer International Publishing, 2020, pp. 383--393.

\bibitem{oehmcke2020creating}
S.~Oehmcke, T.-H.~K. Chen, A.~V. Prishchepov, and F.~Gieseke, ``Creating
  cloud-free satellite imagery from image time series with deep learning,'' in
  \emph{Proceedings of the 9th ACM SIGSPATIAL International Workshop on
  Analytics for Big Geospatial Data}, 2020, pp. 1--10.

\bibitem{chen2019thick}
Y.~Chen, L.~Tang, X.~Yang, R.~Fan, M.~Bilal, and Q.~Li, ``Thick clouds removal
  from multitemporal zy-3 satellite images using deep learning,'' \emph{IEEE
  Journal of Selected Topics in Applied Earth Observations and Remote Sensing},
  vol.~13, pp. 143--153, 2019.

\bibitem{ebel2020multisensor}
P.~Ebel, A.~Meraner, M.~Schmitt, and X.~X. Zhu, ``Multisensor data fusion for
  cloud removal in global and all-season sentinel-2 imagery,'' \emph{IEEE
  Transactions on Geoscience and Remote Sensing}, vol.~59, no.~7, pp.
  5866--5878, 2020.

\bibitem{borkiewicz2021cloudfindr}
K.~Borkiewicz, V.~Shah, J.~P. Naiman, C.~Shen, S.~Levy, and J.~Carpenter,
  ``Cloudfindr: A deep learning cloud artifact masker for satellite dem data,''
  in \emph{2021 IEEE Visualization Conference (VIS)}.\hskip 1em plus 0.5em
  minus 0.4em\relax IEEE, 2021, pp. 1--5.

\bibitem{saxena2022deep}
J.~Saxena, A.~Jain, and P.~R. Krishna, ``Deep learning for satellite image
  reconstruction,'' in \emph{Proceedings of the International Conference on
  Paradigms of Communication, Computing and Data Sciences}.\hskip 1em plus
  0.5em minus 0.4em\relax Springer, 2022, pp. 569--577.

\bibitem{langer2020formatting}
T.~Langer, N.~Fedorova, and R.~Hagensieker, ``Formatting the landscape: Spatial
  conditional gan for varying population in satellite imagery,'' \emph{arXiv
  preprint arXiv:2101.05069}, 2020.

\bibitem{de2019unsupervised}
K.~L. de~Jong and A.~S. Bosman, ``Unsupervised change detection in satellite
  images using convolutional neural networks,'' in \emph{2019 International
  joint conference on neural networks (IJCNN)}.\hskip 1em plus 0.5em minus
  0.4em\relax IEEE, 2019, pp. 1--8.

\bibitem{lossfunction}
\BIBentryALTinterwordspacing
N.~Abraham and N.~M. Khan, ``A novel focal tversky loss function with improved
  attention u-net for lesion segmentation,'' 2018. [Online]. Available:
  \url{https://arxiv.org/abs/1810.07842}
\BIBentrySTDinterwordspacing

\bibitem{lin2019remote}
D.~Lin, G.~Xu, X.~Wang, Y.~Wang, X.~Sun, and K.~Fu, ``A remote sensing image
  dataset for cloud removal,'' \emph{arXiv preprint arXiv:1901.00600}, 2019.

\end{thebibliography}

\end{document}